\documentclass[sigconf, nonacm]{acmart}

\usepackage{xspace}
\usepackage{balance}
\usepackage{subfig}

\usepackage{multirow} 
\usepackage{enumitem}

\DeclareMathOperator*{\argmax}{arg\,max}
\usepackage{pifont}

\newcommand{\xmark}{\ding{55}}%

\newcommand{\eg}{\emph{e.g.,}\xspace}

\setcopyright{acmcopyright}
\copyrightyear{2023}
\acmYear{2023}
\setcopyright{acmlicensed}
\acmConference[SIGIR '23]{Proceedings of the 46th International ACM SIGIR Conference on Research and Development in Information Retrieval}{July 23--27, 2023}{Taipei, Taiwan}
\acmBooktitle{Proceedings of the 46th International ACM SIGIR Conference on Research and Development in Information Retrieval (SIGIR '23), July 23--27, 2023, Taipei, Taiwan}
\acmPrice{15.00}
\acmDOI{10.1145/3539618.3591839}
\acmISBN{978-1-4503-9408-6/23/07}

\acmSubmissionID{3896}
\begin{document}

\title{Context-Aware Classification of Legal Document Pages}

\author{Pavlos Fragkogiannis}
\orcid{0009-0001-0456-2743}
\affiliation{%
  \institution{Thomson Reuters Labs}
  \streetaddress{5 Canada Square, Canary Wharf}
  \city{London}
  \country{UK}
  \postcode{E14 5AQ}
}
\email{pavlos.fragkogiannis@thomsonreuters.com}

\author{Martina Forster}
\orcid{0009-0007-5521-091X}
\affiliation{%
  \institution{Thomson Reuters Labs}
  \streetaddress{Landis + Gyr-Strasse 3}
  \city{Zug}
  \country{Switzerland}
  \postcode{6300}
}
\email{martina.forster@thomsonreuters.com}

\author{Grace E. Lee}
\orcid{0000-0001-7458-414X}
\affiliation{%
  \institution{Thomson Reuters Labs}
  \streetaddress{5 Canada Square, Canary Wharf}
  \city{London}
  \country{UK}
  \postcode{E14 5AQ}
}
\email{grace.lee2@thomsonreuters.com}

\author{Dell Zhang}
\orcid{0000-0002-8774-3725}
\affiliation{%
  \institution{Thomson Reuters Labs}
  \streetaddress{5 Canada Square, Canary Wharf}
  \city{London}
  \country{UK}
  \postcode{E14 5AQ}
}
\email{dell.z@ieee.org}

\renewcommand{\shortauthors}{Fragkogiannis et al.}

\begin{abstract}
For many business applications that require the processing, indexing, and retrieval of professional documents such as legal briefs (in PDF format etc.), it is often essential to classify the pages of any given document into their corresponding types beforehand.
Most existing studies in the field of document image classification either focus on single-page documents or treat multiple pages in a document independently. 
Although in recent years a few techniques have been proposed to exploit the context information from neighboring pages to enhance document page classification, they typically cannot be utilized with large pre-trained language models due to the constraint on input length.
In this paper, we present a simple but effective approach that overcomes the above limitation. 
Specifically, we enhance the input with extra tokens carrying sequential information about previous pages --- introducing recurrence --- which enables the usage of pre-trained Transformer models like BERT for context-aware page classification. 
Our experiments conducted on two legal datasets in English and Portuguese respectively show that the proposed approach can significantly improve the performance of document page classification compared to the non-recurrent setup as well as the other context-aware baselines.
\end{abstract}

\maketitle

\section{Introduction}

Automatically analyzing business documents such as business invoices, court briefs or tax forms is an important task in the legal, medical and tax domains. However, it is also incredibly challenging, due to the diversity of format, layout, and specialized contents. 

In an early stage of document processing, a page type determines content types which need to be extracted and analyzed. Depending on content types, different extraction and analysis techniques may be applied. Formally, the problem setting is: given a document consisting of multiple pages, the task is to assign a type to each page of a document from a predefined set of types. 

In theory, the task could be optimally solved by combining large pre-trained models (\eg LayoutLM~\cite{huang_layoutlmv3_2022}) with contextual information from neighboring pages. In practice, combining these two is not as straightforward or practical, as these models have a max token-limitation constraint and are normally limited to classifying a single page at a time.
Linear-chain Conditional Random Fields (CRF)~\cite{lafferty_conditional_2001} have been leveraged in the literature as an alternative to jointly predict pages of a document~\cite{luz_de_araujo_victor_2020, luz_de_araujo_sequence-aware_2022}. These are, however, hard to combine with a pre-trained language model outside of the sequence/token classification setup. Sequence classification models (\eg LSTM~\cite{hochreiter_long_1997}) face similar challenges.

We propose a simple approach for the task which combines the best of both worlds, and evaluate its performance compared to several baselines. Our model leverages the sequential order of pages and classifies a given page utilizing the previous page. Specifically, we combine a BERT~\cite{devlin_bert_2019} model with the encoding of the previous page type and perform each page type prediction based not only on its page content but also on previous page types. We validate the model performance using public and private datasets and results are consistent; this simple model outperforms baselines including multi-modal models. Furthermore, our approach advances state-of-the-art results for SVic+~\cite{luz_de_araujo_sequence-aware_2022} by 3.49\% macro-average $F_1$ score.

Our proposed approach extends out-of-the-box pre-trained models and is easy to implement as it does not involve modifications in the model architecture, rather, the input to it. This lightweight model modification can be easily plugged into and enhance the document processing pipeline for various industry applications.

\begin{figure*}[ht]
\centering
\includegraphics[scale=0.24]{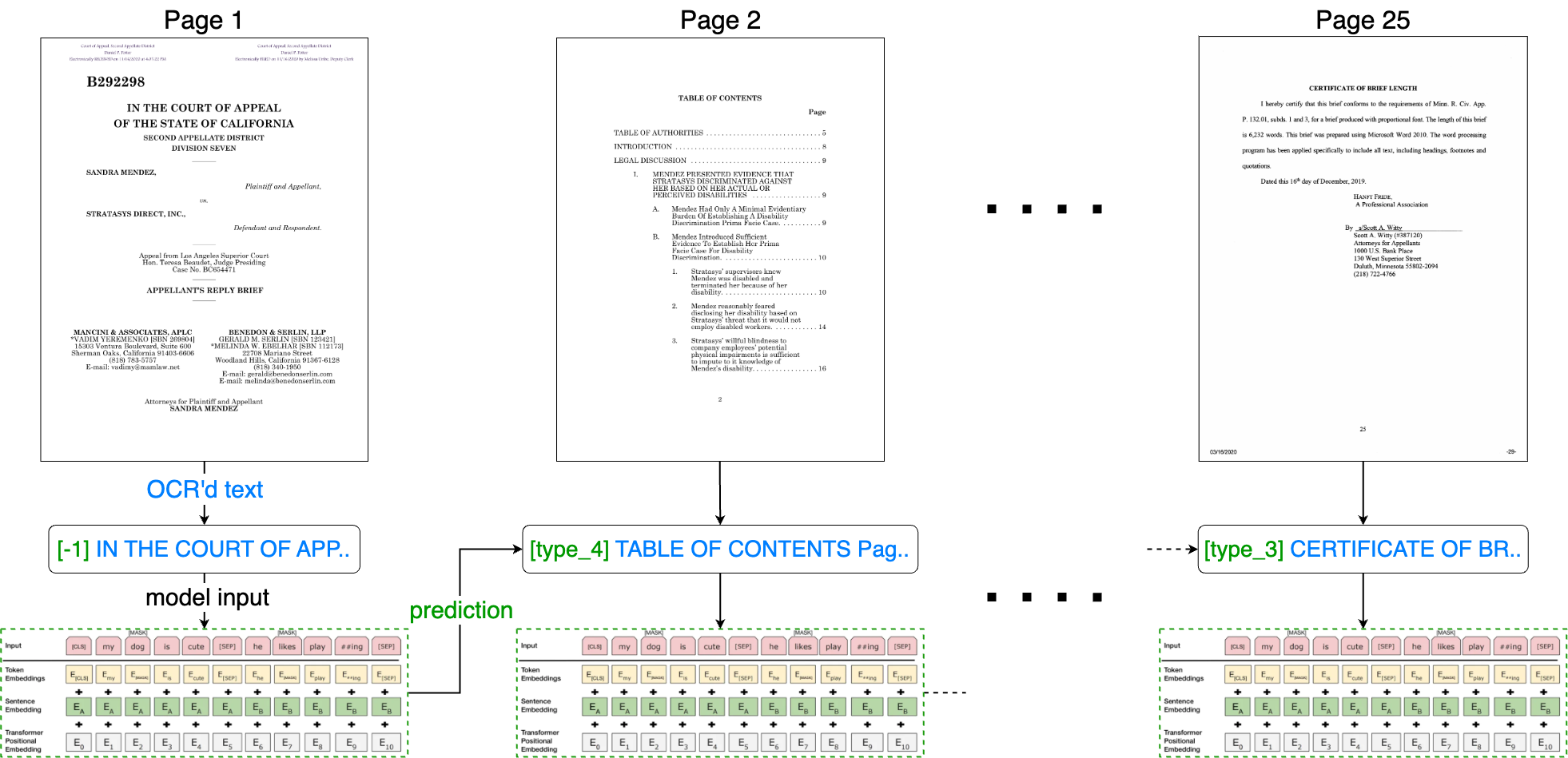}
\caption{Our recurrent BERT architecture in action; special tokens corresponding to the previous page prediction(s) are prepended to each page's input string. The $[-1]$ token is reserved for the first page of a document.}
\label{fig:recurrent_BERT}
\end{figure*}

\section{Related Work}




\subsection{Document Image Classification (DIC)}
Early work on image-based classification~\cite{harley_evaluation_2015, das_document_2018} was further advanced with the use of additional modalities in the input~\cite{dauphinee_modular_2019}. 
The emergence of pre-trained Transformer models~\cite{devlin_bert_2019} led to strong image and text fusion models~\cite{bakkali_visual_2020} and, more recently, state-of-the-art methods have been focusing on multi-modal models that additionally incorporate layout information in the form of positional bounding boxes of tokens, such as LayoutLM~\cite{huang_layoutlmv3_2022} and DocFormer~\cite{appalaraju_docformer_2021}.

Not many English language datasets have been made public for experimentation on the DIC task, with the majority of the literature focusing on tobacco litigation documents datasets~\cite{kumar_structural_2014, harley_evaluation_2015, biten_ocr-idl_2022}. 
Limited by the datasets' availability, existing work has not focused on the task of classifying pages as part of the same document, which requires leveraging different aspects of information apart from multiple modalities, such as page arrangement and inter-dependencies. \\

\subsection{Memory in Transformers}
This is a branch of the Memory-Augmented Neural Networks research, aiming to improve the memory capacity of the Transformer model. One line of research focuses on reforming the existing model architecture, from reusing hidden state representations as a form of segment-level recurrence to incorporating external memory components 
\cite{dai_transformer-xl_2019, al_adel_memory_2021}. Our approach is closer to another line of research, which aims to keep the model architecture intact and instead enhance the input representation with special tokens. 
Chan et al.~\cite{chan_recurrent_2019} insert special $[HL]$ tokens to indicate the answer phase to the question generation model, which tackles ambiguities in long inputs. Burtsev et al.~\cite{burtsev_memory_2021} prepend 
$[mem]$ tokens to the input sequence, as a placeholder that the model uses to store and process global or local representations. 

To the best of our knowledge, this is the first work that considers input-modified memory-augmented Transformer models for the problem of document page classification.

\section{Approaches}

\subsection{Baseline Method} \label{sec:baseline}

In a context-oblivious page-level approach, an input is a single page and no concept of documents is considered. An advantage of this approach is that large pre-trained models can be fine-tuned, so less training data is required. However, this can only work for tasks where the information of the page itself is sufficient to classify it, as no sequence information can be leveraged. 

Assuming a multi-class classification task of a text-only neural network model ($NN$) with a set of learned parameters $\theta$ for simplicity and a document with $l$ pages, where $x_1^t, x_2^t, .., x_{m_t}^t$ are the input tokens for the $t$-th page and there are $n$ possible page-type classes, the model is trained to predict a score $y_i^t$ for each class:
\begin{equation}
\label{eq:simple_classification}
    y_i^{t} = NN_{\theta}\left(x_{1:{m_t}}^{t}\right) \qquad \text{for } 1 \leq t \leq l \ .
\end{equation}
Then the predicted class for the $t$-page of the given document is:
\begin{equation}
\label{eq:predicted_class}
  C^t = \argmax_i{y_i^{t}} \;\;\qquad \text{for } 1 \leq t \leq l \ .
\end{equation}

\subsection{Proposed Method}




The lack of context of the 
page-level approach can be a breaking factor for our task in terms of performance. In order to enhance our pre-trained BERT model, we leverage page type special tokens as additional feature input to the model, which the model is supplying to itself on every iteration. Applied to the multi-class classification task with $n$ possible page types from Section~\ref{sec:baseline}, we modify the input text before tokenization such that a special token corresponding to the previous page predicted class is prepended to the input text, among possible [type\_1], [type\_2], …, [type\_$n$] tokens. Figure~\ref{fig:recurrent_BERT} shows an inference example.



Expanding Eq. \eqref{eq:simple_classification}, we modify the problem formulation such that the classification is performed in a recurrent manner, where each page's classification score is also conditioned on the prediction of the previous page, out of a total $l$ pages in a document:
\begin{equation}
\label{eq:conditional_classification}
    y_i^{t}  = \left\{
    \begin{array}{ll}
      NN_{\theta}\left(x_{1:{m_t}}^{t}\right)  & \text{for } t = 1 \ , \\ 
      NN_{\theta}\left(C^{t-1}, x_{1:{m_t}}^{t}\right) & \text{for } 1 < t \leq l \ . 
    \end{array} \right.
\end{equation}
Under this setup, the model performs inference iteratively in a document one page at a time in a recurrent fashion, incorporating its predictions for a given page in the input of the next page text input. The proposed method can be expanded to multi-label classification setups by prepending multiple page type tokens to the input string, which is the case with our US Appellate Briefs dataset.



\section{Evaluation}

\subsection{Datasets}
\label{sec:datasets}


\textbf{US Appellate Briefs} is our proprietary dataset consisting of 781 documents of 22,143 pages and originating from 52 jurisdictions across the US, annotated with 7 page types by subject matter experts at Thomson Reuters (see Table~\ref{tab:class_counts}). Each page can have one or more labels, so it is a multi-label classification dataset.

\textbf{SVic+}~\cite{luz_de_araujo_sequence-aware_2022} is a public dataset in Portuguese that includes 6,510 lawsuits from the Brazilian Supreme Court, consisting of 94,267 documents of 339,478 pages in total, annotated with 6 document types. Only a single label is used for every page, hence it's a multi-class classification dataset. 
For our experiments, we consider each lawsuit as a sequence of pages, and thus classify document pages in the context of their corresponding lawsuit.

\subsection{Experimental Setup}


\paragraph{\textbf{Baselines}} For context-oblivious methods, we experiment with state-of-the-art uni-modal and multi-modal pre-trained language models, specifically BERT, BERTimbau~\cite{souza_bertimbau_2020} (text) and LayoutLMv3 (text + image + layout). As context-aware method, we use a BiLSTM model where the input is the encoded representation of each page in a document, which we obtain using TF-IDF vectors (with a vocabulary size of 60,000 words for the US briefs dataset) and then reduce to 300 dimensions using Singular Value Decomposition (SVD)~\cite{klema_singular_1980}.

For SVic+, we also compare our performance to the numbers reported in the original paper~\cite{luz_de_araujo_sequence-aware_2022} for the following methods: LayoutXLM\cite{xu_layoutxlm_2021} (multi-lingual version of LayoutLMv2 model), FM (early fusion model of text and image embeddings), FM+CRF (conditional random field enhanced version of FM), and BiLSTM-F (BiLSTM model with concatenated image and text embeddings as input).

In order to validate the effectiveness of our approach, we additionally experiment with CRF as an alternative method to incorporate context to our context-oblivious BERTimbau model. CRF is typically used on sequence classification setups, for example an LSTM or a BERT token classification or NER model~\cite{huang_bidirectional_2015, ma_end--end_2016, souza_portuguese_2020}, as a layer that is added on top of the overall model architecture and is fine-tuned along with the base model. However, for our page-level BERTimbau model, we are limited to training the CRF layer separately by using the saved predictions of the BERTimbau model and training the CRF with those, essentially using a frozen fine-tuned model as a feature extractor, similarly to Araujo et al.~\cite{luz_de_araujo_sequence-aware_2022}.


\paragraph{\textbf{Reproducibility}} 

For reproducibility of our results on the public dataset, SVic+, we report our experimental settings with the BERTimbau (base) model, which we load and train using HuggingFace and its Trainer API\footnote{https://huggingface.co/docs/transformers/main\_classes/trainer}. Both the original and the recurrent version of the model are fine-tuned for 6 epochs, using a batch size of 32. We apply the default optimization settings of the Trainer API, specifically optimizing the cross-entropy loss using AdamW~\cite{loshchilov_decoupled_2019} optimizer and a linear learning rate scheduler, with initial learning rate of $2e-5$ and warmup steps accounting to 10\% of the total training steps. We use mixed precision (FP16) to reduce the memory usage of the neural network model during fine-tuning. Regarding preprocessing, we use the SVic+ dataset as-is, since it is provided in an already preprocessed format.


\paragraph{\textbf{Teacher Forcing}}

For simplicity and efficiency reasons, we perform the fine-tuning of the recurrent model using teacher forcing~\cite{williams_learning_1989}, where the \textit{ground truth labels} of the previous page are prepended in the input instead of the actual model predictions. This allows us to perform batched training which offers considerable efficiency gains during training, using the same configuration as the original model. Nevertheless, inference during the evaluation phase is performed sequentially, page after page, and the model supplies the predictions of the previous page to itself.

\begin{table}[tb]\small
  \caption{Number of pages in our US Appellate Briefs dataset. }
  \label{tab:class_counts}
  \begin{tabular}{l|rrr}
    \toprule
    \textbf{Class}          & \textbf{Training set} & \textbf{Validation set} & \textbf{Test set}\\
    \midrule
    Caption                 &   772 &   90 &  103 \\
    Table of Contents       &   858 &  102 &  104 \\
    Table of Authorities    &  1331 &  173 &  159 \\
    Document Body           & 13956 & 1627 & 1689 \\
    Signature               &   673 &   75 &   85 \\
    Certificate of Service  &   401 &   45 &   48 \\
    Other Certificate       &   589 &   73 &   82 \\
    \bottomrule
  \end{tabular}
\vspace{-4mm}
\end{table}



\begin{table*}[ht]\small
  \caption{Test set $F_1$ scores per class on the US appellate briefs dataset of our approach, along with various uni-modal and multi-modal baselines. Compared to the original BERT model, our approach achieves significant improvement on specific classes that benefit the most from context-awareness, such as +8.47\% $F_1$ on \textit{Other Certificate} and +6.72\% $F_1$ on \textit{Signature} pages.
  }
  \label{tab:brief_scores}
  \begin{tabular}{lc|ccccccc|cc}
    \toprule
    \textbf{Method} & \textbf{Context} & \textbf{Caption} & \textbf{Cert. of Service} & \textbf{Doc. Body} & \textbf{Other Cert.} & \textbf{Sign.} & \textbf{ToA} & \textbf{ToC} & \textbf{macro-avg} & \textbf{weighted-avg} \\
    \midrule
    BiLSTM & \checkmark & 87.63 & 86.00 & 98.78 & 80.00 & 91.23 & 95.90 & 91.71 & 90.18 & 96.45\\
    LayoutLMv3 & \xmark & 96.65 & 94.95 & 99.64 & 91.50 & 92.68 & \textbf{98.76} & 98.56 & 96.11 & 98.75\\
    BERT & \xmark & 93.72 & 95.83 & 99.58 & 88.46 & 89.14 & 98.14 & 98.10 & 94.71 & 98.28\\
    \textbf{BERT+recurrence} & \checkmark & \textbf{98.06} & \textbf{97.96} & \textbf{99.67} & \textbf{96.93} & \textbf{95.86} & 98.75 & \textbf{98.58} & \textbf{97.97} & \textbf{99.21}\\
    \bottomrule
  \end{tabular}
\end{table*}

\begin{table*}[tb]\small
  \caption{Test set $F_1$ scores per class on the SVic+ dataset of our approach, compared to different methods reported in the literature. Our method achieves better performance on 4 out of 6 classes on the task, as well as higher macro-average and weighted-average $F_1$ scores.}
  \label{tab:svic_scores}
  \begin{tabular}{lc|cccccc|cc}
    \toprule
    \textbf{Method} & \textbf{Context} & \textbf{Acórdão} & \textbf{ARE} & \textbf{Despacho} & \textbf{Others} & \textbf{RE} & \textbf{Sentença} & \textbf{macro-avg} & \textbf{weighted-avg}\\
    \midrule
    LayoutXLM~\cite{luz_de_araujo_sequence-aware_2022} & \xmark & 60.13 & 23.91 & 23.72 & 96.54 & 72.25 & 66.68 & 57.21 & 92.81 \\
    FM~\cite{luz_de_araujo_sequence-aware_2022} & \xmark & 90.74 & 57.92 & 63.98 & 97.24 & 75.47 & 82.04 & 77.90 & 94.72 \\
    FM+CRF~\cite{luz_de_araujo_sequence-aware_2022} & \checkmark & \textbf{91.56} & 60.74 & 62.69 & 97.67 & 78.43 & 83.42 & 79.09 & 95.38 \\
    BiLSTM-F~\cite{luz_de_araujo_sequence-aware_2022} & \checkmark & 88.97 & 61.16 & 64.07 & 97.46 & 79.67 & 85.26 & 79.43 & 95.30 \\
    BERTimbau & \xmark & 82.31 & 62.78 & 65.71 & 97.33 & 77.02 & 80.46 & 77.60 & 94.95 \\
    BERTimbau+CRF & \checkmark & 84.99 & 67.05 & \textbf{66.67} & 97.70 & 78.55 & 83.54 & 79.75 & 95.52 \\
    \textbf{BERTimbau+recurrence} & \checkmark & 90.71 & \textbf{73.42} & 64.33 & \textbf{97.89} & \textbf{83.54} & \textbf{87.63} & \textbf{82.92} & \textbf{96.22} \\
    \bottomrule
  \end{tabular}
\end{table*}


\subsection{Results and Discussion}

\paragraph*{\textbf{US Appellate Briefs}}
Experimental results in Table~\ref{tab:brief_scores} confirm our core hypothesis; while the multi-modal LayoutLMv3 can indeed take advantage of the additional modalities and perform better than the regular text-only BERT model, the additional context that the recurrence approach offers can be far more important in some cases. Our context-aware method improves $F_1$ scores on almost all classes except one compared to LayoutLMv3, despite having 2 modalities less to utilize. On the contrary, the context-aware BiLSTM model fails to perform better than any context-oblivious model in most classes, which could be due to the relatively small size of the dataset where the pre-trained models can be more efficient. \\


\paragraph*{\textbf{SVic+}}
As presented in  Table~\ref{tab:svic_scores}, our recurrent BERTimbau method achieves the highest performance than the other models on 4 classes in terms of $F_1$ score, as well as +3.49\% macro average and +0.84\% weighted average $F_1$ scores compared to previous state-of-the-art. To understand the result better, we calculated page class statistics and it shows that for the average class, the probability of encountering the same class on the next page is 70.62\%. Interestingly, the corresponding probability score for the class we have the highest $F_1$ score increase, ARE, is 89.14\%, while the ones for the two classes we do not improve on, Acórdão and Despacho, are 28.21\% and 31.82\%, respectively. In fact, the median number of pages for those two document types is 1 (see Table \ref{tab:median_doc_length_svic}). Our understanding is that the proposed approach works best for classes that are typically longer than one page.

\begin{table}[tbh]\small
  \caption{Document statistics of test set of the SVic+ dataset.}
  \label{tab:median_doc_length_svic}
  \begin{tabular}{l|ccc}
    \toprule
    \textbf{Document Class} & \textbf{Median length} & \textbf{Max length} & \textbf{\#Total pages} \\
    \midrule
    Acórdão & 1 & 5 & 273 \\
    ARE & 4 & 64 & 1,841 \\
    Despacho & 1 & 10 & 198 \\
    Others & 11 & 1,044 & 85,408 \\
    RE & 19 & 107 & 6,331 \\
    Sentença & 5 & 42 & 1,475 \\
    \bottomrule
  \end{tabular}
\end{table}

The benefit of using the proposed page type special tokens is clearly shown by comparing with the original BERTimbau model: The recurrent version achieves an average +8.18\% $F_1$ score increase on Acórdão, ARE, RE and Sentença classes. Interestingly, it also gets a -1.38\% $F_1$ score decrease on the Despacho class, on which the probability of encountering the same class on the following page is 31.82\%. This is in line with our understanding from before.

Compared to the BERTimbau+CRF version, our recurrent BERTimbau model achieves a +3.17\% macro average $F_1$ score increase, which shows the benefits of modelling sequential dependencies on the feature level as opposed to the prediction level. Nevertheless, the CRF version does provide notable performance increase compared to the original model across all classes, but still only a slight improvement over the previous state-of-the-art.

The performance improvements of the recurrent BERTimbau model over the original and CRF versions are \emph{statistically significant} ($p\text{-value}\ll0.01$) according to the McNemar-Bowker test~\cite{bowker_test_1948}, a generalization of McNemar's test~\cite{mcnemar_note_1947} to multi-class classification setups.

\section{Conclusion}
In this work, we present a simple approach to applying large pre-trained models to sequence-aware page classification. Despite the input length limitation for Transformer models such as BERT, we present a way to leverage information from the previous page, making the model aware of the context within a document. Our results show that our approach can significantly improve performance compared to the non-recurrent setup and alternative context-aware approaches, while being simple and straightforward. These findings can be beneficial in particular in low-resource settings of page classification, where transfer learning decreases the need for large amounts of training data, but the document context can still be leveraged at classification time. We leave experimentation of our special token method with multi-modal models for future work.


\section*{Company Portrait}
\textbf{Thomson Reuters}\footnote{\url{https://www.thomsonreuters.com/en.html}} is best known for the globally respected Reuters News agency, but our company is also the leading source of information for legal, corporate, and tax \& accounting professionals.
We have over 60,000 TBs worth of legal, regulatory, news, and tax data.
In 2022, Thomson Reuters Labs celebrated its 30-year anniversary, a testament to the team's history of AI ingenuity.

\section*{Presenter Bio}

\textbf{Pavlos Fragkogiannis} is an Applied Research Scientist at Thomson Reuters Labs in London, UK. He has a Master's degree in Computer Science from Imperial College London. His research focuses on Information Retrieval and Natural Language Processing.



\newpage

\bibliographystyle{unsrt}
\balance
\bibliography{anthology,custom}

\end{document}